\documentclass{article}

\usepackage[preprint]{neurips_2019}

\usepackage[utf8]{inputenc} 
\usepackage[T1]{fontenc}    
\usepackage{hyperref}       
\usepackage{url}            
\usepackage{booktabs}       
\usepackage{amsfonts}       
\usepackage{nicefrac}       
\usepackage{microtype}      
\usepackage{graphicx}       
\graphicspath{ {./} }
\usepackage{color}
\usepackage[dvipsnames]{xcolor}

\usepackage{amsmath, amsfonts,amsthm,amssymb}       
\usepackage{xspace}
\newcommand{\xxxVAE}{{SimVAE}\xspace}
\newcommand{\betaVAE}{{$\beta$-VAE}\xspace}

\title{SimVAE: Simulator-Assisted Training for Interpretable Generative Models}

\author{%
  Akash Srivastava\thanks{Equal Contribution.} \\
  MIT-IBM Watson AI Lab\\
  IBM Research\\
  Cambridge, MA \\
  \texttt{akash.srivastava@ibm.com} \\
   \And
   Jessie Rosenberg$^*$ \\
  MIT-IBM Watson AI Lab\\
  IBM Research\\
  Cambridge, MA \\
  \texttt{jcrosenb@us.ibm.com} \\
   \AND
  Dan Gutfreund \\
  IBM Research\\
  Cambridge, MA \\
  \texttt{dgutfre@us.ibm.com} \\
   \And
  David D. Cox \\
  MIT-IBM Watson AI Lab\\
  IBM Research\\
  Cambridge, MA \\
  \texttt{david.d.cox@ibm.com} \\
}

\begin{document}

\maketitle

\begin{abstract}
This paper presents a simulator-assisted training method (\xxxVAE) for variational autoencoders (VAE) that leads to a disentangled and interpretable latent space. Training \xxxVAE is a two step process in which first a deep generator network (decoder) is trained to approximate the simulator. During this step the simulator acts as the data source or as a teacher network. Then an inference network (encoder) is trained to invert the decoder. As such, upon complete training, the encoder represents an approximately inverted simulator. By decoupling the training of the encoder and decoder we bypass some of the difficulties that arise in training generative models such as VAEs and generative adversarial networks (GANs). 
We show applications of our approach in a variety of domains such as circuit design, graphics de-rendering and other natural science problems that involve inference via simulation. 
\end{abstract}

\section{Introduction}

Simulation as a scientific tool is as old as scientific exploration itself. From the ancient Greeks who drew circles in the sand to discover the connection between radius and circumference, to modern simulations of complex atomic reactions, protein folding and photo-realistic computer graphics, simulators represent human knowledge in a well-defined symbolic form, crystallizing information into models that can generate output data based on particular input specifications. Much of our progress in understanding the world relies on developing simulators, often using several of them in concert to describe larger interconnected systems. 

Humans (and animals in general) also distill information from the world, but rather than explicitly knowing and manipulating precise equations governing e.g. the laws of physics, they have an intuitive sense of physics that allows split second, ``good enough'' estimates; for instance, even a dog can catch a ball mid-air without explicitly knowing the laws of physics or manipulating equations.
Therefore, one might speculate that our brains possess some form of a ``simulation engine in the head,'' which distills knowledge of the world into simpler heuristics that help us in our everyday lives \citep{Lake:2016}. Moreover, a desirable feature of such a simplified model would be to allow inference in both the forward and inverse directions of the simulator, in contrast to traditional numerical simulators which are difficult or impossible to invert.


In this paper we address the question of how to distill symbolic mappings modeled by complex and not necessarily differentiable simulators into neural networks.
Classification systems can learn a map between data points and labels, but there is no continuity in the output space and these systems often cannot generalize well to new types of examples. Generative models understand their domain well enough to create new examples, they possess smooth mappings over the input-output space, and can operate probabilistically. However, the latent space typically does not correspond to any human-understandable set of parameters; therefore it is difficult to generate output data with a specified set of features. If a larger system needs to take input from multiple domains, all component parts must be trained together end-to-end, since there is no interpretable and standard interface between components that would allow different units to be swapped out or combined in different configurations. In recent years, a few approaches to address these issues have been proposed \citep{infogan,bvae,kulkarni2015deep}. In Section \ref{sec:related} we compare our approach to these works.  



Here we present \xxxVAE, a method for training a variational autoencoder model of a simulator, resulting in both a generator that represents a simplified, probabilistic version of the simulator, and an encoder that is a corresponding probabilistic inverse of the simulator. Though simulators can in general be quite complex and may be discontinuous, a heuristic, smooth version that is invertible can be valuable for applications in downstream reasoning tasks, or in technical fields such as circuit design, protein folding, and materials design, among others. Such a simplified simulator can also be used as a guide for further detailed inquiry into specific parameter spaces of interest that could take place back in the original, more accurate simulation domain. 

The \xxxVAE training method naturally restricts the latent space to be both interpretable and disentangled. Note that this does not require orthogonality, and in fact we show example cases where the natural and interpretable parameters of the simulator are not orthogonal. In this case the \xxxVAE latent space matches the simulator input space rather than artificially constraining its parameters to be orthogonal at the cost of reducing accuracy. \xxxVAE can be trained either in a semi-supervised manner using an explicit simulator, which results in a differentiable, probabilistic model and inverse model of the simulator, or can be trained fully unsupervised by taking the inputs from, for example, a GAN \citep{gan} or InfoGAN \citep{infogan}.

We demonstrate the general applicability of our approach by plugging in several simulators from very different domains such as computer graphics, circuit design and mathematics. This is in contrast to most previous works on learning disentangled representations which focused on understanding images and scenes.




\section{Background}
\label{sec:related}

Variational autoencoders (VAE) \citep{kingma2013auto} and generative adversarial networks (GAN)  \citep{gan} are two of the most popular state-of-art methods for learning deep generative models. Both methods are unsupervised and only need samples $\{x_i\}_{i=1}^M$ from the data distribution $p_x$ to learn a parametric model $G_\phi$ (generator) whose distribution $q_\phi$ matches that of the true data $p_x$. The distributions are matched using discrepancy measures such as $f$-divergences \citep{kullback1951information} or integral probability metrics \citep{mmd}. 


In VAEs, $G_\phi$ represents a probabilistic function that maps sets of samples from the prior distribution $p_z$ over the latent space $\mathbb{R}^K$ to sets of samples in the observation space $\mathbb{R}^N$. In doing so, it additionally requires an encoding function (encoder) or an inference network $E_\theta$ to parameterise the variational posterior $p_\theta(z|x)$, which is trained to approximately invert the mapping of $G_\phi$. VAEs make an assumption about the distribution of the data, which yields a likelihood function. As such they can be trained using stochastic gradient based variational inference. In practice, the VAE training objective is a lower bound to the log-likelihood, also referred to as the ELBO:
\begin{align}
    \label{eq:elbo_0}
    \int_{\mathcal{X}} p_x(x)\log q_\phi(x) dx \geq \int_{\mathcal{X}} p_x(x) \big [ \int_{\mathcal{Z}} p_\theta(z|x) \log \frac{p_z}{p_\theta(z|x)}dz  + \int_{\mathcal{Z}} p_\theta(z|x) \log q_\phi(x|z)dz \big ] dx. 
\end{align}

This ELBO has a unique form, the first term is in fact a negative Kullback–Leibler (KL) divergence between the variational posterior and the model prior over the latent space.
The second term promotes the likelihood of the observed data under the assumed model $\phi$ but it is not always tractable unlike the first term and is usually estimated using Monte Carlo (MC) estimators. 
But MC estimator based training increases the variance of the gradients. This increase in variance is controlled using the \textit{re-parameterisation trick} which, in effect, removes the MC sampler from the computation graph. For example, using this trick $u \sim \mathcal{N}(\mu, \sigma)$ can be expressed as $u = \mu + \sigma*\epsilon$ where $\epsilon \sim \mathcal{N}(0,1)$. This gives the final ELBO
\begin{align}
    \label{eq:elbo}
    \mathbb{E}_{p_x}[\log q_\phi(x)] \geq \mathbb{E}_{p_x} \bigg [ -KL[p_\theta(z|x)\Vert p_z]  + \mathbb{E}_{\mathcal{N}(0,I)} [\log q_\phi(x|\mu_\theta(x) + \sigma_\theta(x)*\epsilon ) ] \bigg]. 
\end{align}
Here $\mu_\theta$ and $\sigma_\theta$ are posterior parameters that the encoder function $E_\theta$ outputs.

\section{Method}
\label{method}


In this section we describe the two-step, simulator-assisted training procedure of \xxxVAE. We formally define the simulator $S: \mathbb{R}^K \mapsto \mathbb{R}^N$ as a  deterministic black-box function that maps each $i$th point $z_i \in \mathbb{R}^K$ from its domain to a unique point $x_i \in \mathbb{R}^N$ in its range. Usually, $N>>K$ for most physical simulators. In the first step, \xxxVAE trains a generator $G_\phi: \mathbb{R}^K \mapsto \mathbb{R}^N$, a Borel measurable function, to learn a probabilistic map of the simulator $S$. In order to achieve this we define $Z(z)$,  in the latent space, and $X(x)$, in the output space, as $K$ and $N$ dimensional random variables. In order to learn the function $S$ faithfully, $G_\phi$ is parameterized with a deep neural network. The training is achieved by minimizing a suitable measure of discrepancy $D$ (depending on the output space of $S$) on the observations from the two functions on the same input with respect to $\phi$, as shown below:
\begin{align}
    \label{eq:decloss}
    \min_\phi \mathcal{L_D}(\phi) = \min_\phi \sum_i D[S(z_i),G_\phi(z_i)].
\end{align}

Since the domain of $S$ is infinite in practice, the optimization problem is solved using mini-batches in a stochastic gradient descent first order optimization method such as ADAM.

Upon successful training of the generator, $G_\phi \approx S$, the next step in \xxxVAE is to train an inference network (encoder, $E_\theta: \mathbb{R}^D \mapsto \mathbb{R}^K$) to invert the generator, which in turn, if successfully trained, will give an approximate inversion of the simulator and as a result a disentangled and interpretable representation of the latent space. We do that via the following objective:


\begin{align}
    \label{eq:encloss}
    \min_\theta \mathcal{L_E}(\theta) = \min_\theta \sum_i D[S(z_i),G_\phi(\mu_{E_\theta}(S(z_i)))].
\end{align}

Here, $\mu_{E_\theta}$ is the posterior parameters that the encoder output. We emphasize that training the encoder does not involve any supervision. Computation is only done on $S(z_i)$, a sample in data space, while the latent variables $z_i$ are hidden. The key point is that while the simulator is typically black-box and non-differentiable, the generator is a neural network which we control and therefore we can backpropagate through the network and the latent variables to train the encoder.
Note, while we use the re-parameterization trick here, in general our method does not require it. If it is not used, in the cases where $S$ is a one-to-many map, the encoder will only learn one such mapping.


\subsection{Comparison to other related approaches}
\label{sec:related}

\paragraph{\xxxVAE vs. Classification.} Having access to a simulator, one could directly train a model of its inverse via the following objective:
\begin{align}
    \label{eq:encpre}
    \min_\theta \mathcal{L_E}(\theta) = \min_\theta \sum_i loss[z_i,E_\theta(S(z_i))],
\end{align}
where $loss$ could be any classification loss (e.g. cross entropy) and/or regression loss (e.g. MSE) depending on $z$'s components. In other words, we could turn the problem into a supervised classification/regression problem. This was done in the past for graphics de-rendering, see for example \citep{de-rendering}.  

There are several reasons why our approach of inverting the generator is preferred. First, $z$ can involve many components which could be very different in nature, turning the problem into a complex multi-task learning problem, where some of the output parameters may be correlated (see the box plotter example in Section \ref{sec:Models}). Furthermore, the inverse of the simulator could potentially be (and often is) a one-to-many relation (see the RLC or polygon examples in Section \ref{sec:Models}), turning the problem into a multi-label learning problem. Most importantly, while loss functions on the latent/symbolic space are useful mathematically and are often used for supervised learning, discrepancies in the observable space, which are the basis for loss functions used to train generative models such as VAE, allow reconstruction of the data distribution and are more natural for humans learning (see Section \ref{sec:discussion}). 



\paragraph{\xxxVAE vs. other disentangled representation learning schemes.}
In the last few years several approaches for learning disentangled representations with generative models have been proposed. Here we survey the ones that are most related to this paper. 

We start with three VAE-based approaches. \betaVAE \citep{bvae} is an unsupervised technique which involves constraining the representations to ensure variable independence at the price of reconstruction accuracy. A hyper-parameter governs that trade-off. We note that the assumption of latent variables independence does not necessarily hold, see the box plots example in Section \ref{sec:Models}. Deep Convolutional Inverse Graphics Network (DC-IGN) \citep{kulkarni2015deep} is a semi-supervised approach for learning interpretable representations of graphics engines. A crucial aspect in training DC-IGN models is the ability to divide the training set into batches in which only one of the latent variables varies (e.g. lighting) while all others are fixed. Training proceeds by clamping all the fixed latent variables to their respective means. We note that while this makes sense for graphics engines, the average of latent variables does not necessarily corresponds to a meaningful data point in the observable space (e.g. see our RLC circuits example). \citep{graphical} incorporate graphical models to the VAE architecture, imposing assumptions on the interpretable variables. Naturally, the specific graphical model depends on the use-case and requires a re-design if the application changes. 


InfoGANs \citep{infogan} learn disentangled representations by regularizing the minimax game between the discriminator and the generator in the GAN framework with an information-theoretic term which aims to maximize the mutual information between a small set of latent variables and the observed data. While InfoGANs have shown impressive results on visual data, still as an unsupervised method it can miss latent information that is important in certain contexts. For example, feeding an image of a function curve to an InfoGAN will likely generate a representation capturing appearance properties of the curve, but it is unlikely that it will retrieve the Fourier coefficients of the function. In addition, InfoGAN, just like other GAN-based approaches, require a significant amount of training while balancing the discriminator and the generator to keep the training stable.

Finally, Wu et. al. \citep{physics} suggest an approach which combines inference models with generative models. However, the models are not trained based on a single simulation but rather independently and thus the approach is not as general as the one we suggest here.

In contrast to the previously mentioned works, our method is general and applies to simulators from a plethora of domains. It does not make any assumptions on the prior or posterior distributions, and it avoids some of the difficulties in jointly training competing (as in GANs) or complementing (as in VAE) models. It relies of course, on the existence of a simulator but as we argue, simulators are everywhere when thinking about them broadly, even the world can be viewed as a simulator (see Section \ref{sec:discussion}).

\section{Experiments}
\label{others}
We trained the \xxxVAE on a variety of simulators, both image-based and purely symbolic, to show the generality of this training method. We used the ADAM \citep{adam} optimiser for training, with the learning rate set to $0.001$ and the other parameters set to their default values in the Tensorflow framework. 
All image-based models were trained using the DCGAN architecture (\cite{radford2015unsupervised}), and the
RLC circuit simulator model was trained using a simple feed-forward architecture. The architecture of the \xxxVAE is interchangeable depending on the desired task, e.g. a RNN could be used for circuit simulation to capture larger signal spaces, or an autoregressive decoder could be used for image generation tasks to enhance reproduction accuracy.

\begin{figure}
  \centering
  \includegraphics[width=0.9\textwidth]{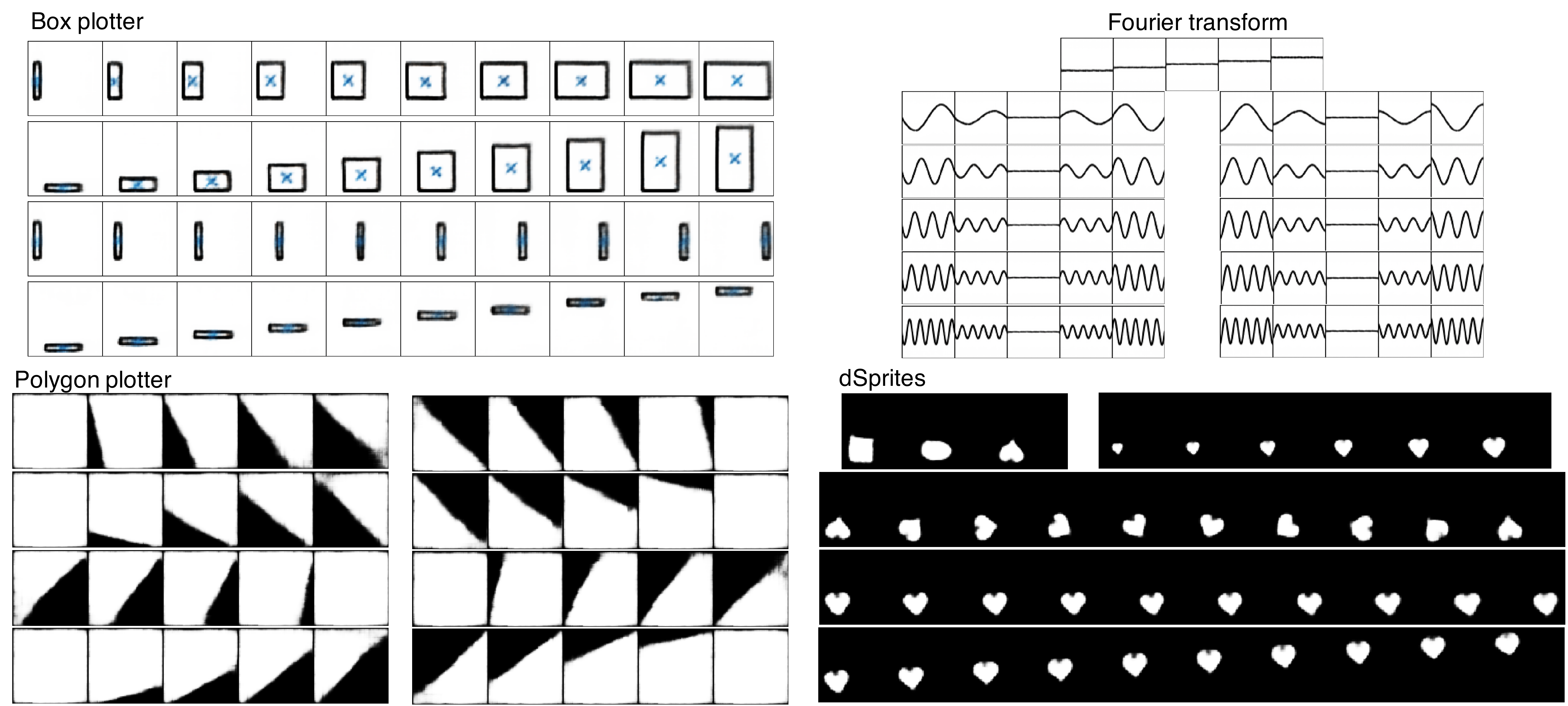}
  \caption{\textbf{Latent transversals.} For each of the four image-based simulators, the output of the generator upon transversing one latent variable and holding the others fixed is shown for each of the latent variables. Box plotter: [width, height, $x$-position, $y$-position], Fourier transform: [constant, 5 cosine coefficients, 5 sine coefficients], Polygon plotter: [($x$,$y$) coordinates for each of four points of the polygon], dSprites: [shape, scale, rotation, $x$-position, $y$-position].
} 
  \label{fig:transversals}
\end{figure}

Comparisons of the output from the model vs. the ground truth simulations are shown in Fig. \ref{fig:output} for all image-based models, and Fig. \ref{fig:RLC} for the circuit model. Representative samples of model output from varying one latent variable while holding the others fixed are shown in Fig. \ref{fig:transversals}. Note that all latent variables in all models are shown, there are no additional latent variables that do not represent interpretable parameters.

\subsection{Trained models}
\label{sec:Models}

\paragraph{Box plotter}
We trained a model on a simple graphical simulator written in Matplotlib (\cite{Hunter:2007}) which plots a black rectangle with a blue 'x' in the center. Input parameters were the $(x,y)$ coordinates of the lower left corner of the rectangle, and the width $w$ and height $h$ of the rectangle.
The sample space that the model was trained on was constrained for the rectangle to remain always within the 64x64 pixel field of view, by constraining the initial $(x,y)$ coordinates based on the randomly sampled $w$ and $h$. This intentionally creates correlations in the ground truth simulator latent variables, which must be replicated in the model latent space for optimum accuracy and interpretability. 

\begin{figure}
  \centering
  \includegraphics[width=0.9\textwidth]{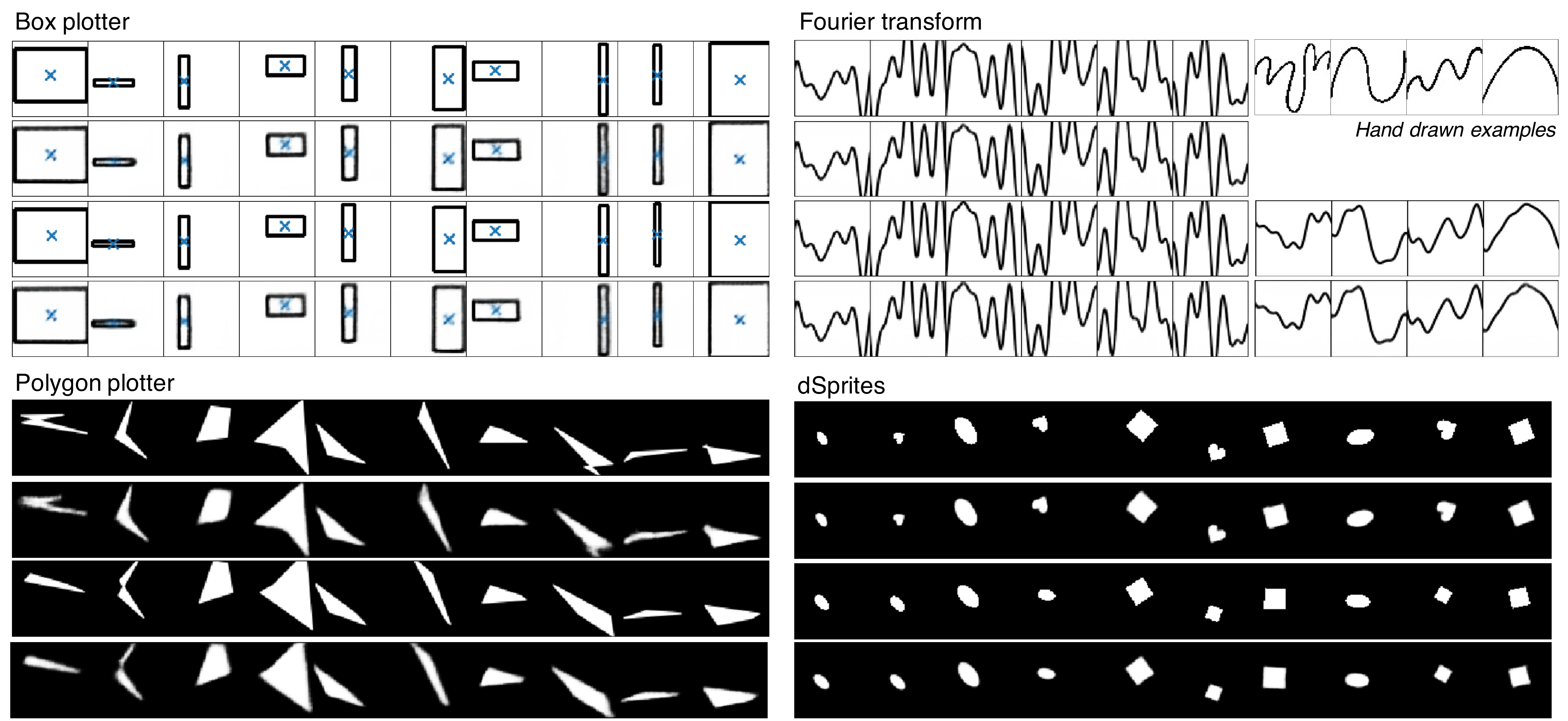}
  \caption{\textbf{Output of simulator, encoder and generator for image-based models.} For each of the four image-based simulators, random samples are shown of the simulator output $X$ on input $Z$, the generator output $\bar{X}$ on the same input $Z$, the simulator output $X$ on encoder output $\bar{Z}$, and the generator output $\bar{X}$ on encoder output $\bar{Z}$, arranged top to bottom in each section. For the Fourier transform, we also show inference on four hand drawn images (right), excepting the generator $\bar{X}$ from input $Z$ as the input $Z$ is not available in this case.}
  \label{fig:output}
\end{figure}

\paragraph{Polygon plotter}
\label{sec:polygon}
This method generalizes straightforwardly to rendering and de-rendering 
an arbitrary four-pointed polygon. The polygon is not restricted to be non-intersecting, and therefore generates folded shapes which the model is also required to learn. Note that the model can choose multiple permutations of the four points of the polygon to generate the same output shape, so the inverse simulator is a one-to-many map. 

\paragraph{dSprites}
We also trained a \xxxVAE on the dSprites disentanglement dataset (\cite{bvae}), consisting of 737,280 binary 2D shapes constructed from five independent generative parameters (shape, size, rotation, $x$-position, and $y$-position). Since the dataset is complete, including all examples of images from the five parameters, we can treat the dataset itself as a black box simulator and invert it using the method described above, although the discrete nature of the shape parameter prevents the VAE from accurately learning this variable within the minimal latent space provided.


\paragraph{Fourier transform}
To demonstrate the advantages of an inverse simulator that operates probabilistically, we train a model based on a simulator which takes as input a set of Fourier coefficients, and outputs a plot of the inverse Fourier transform based on those coefficients. In addition to the performance of the model on input from the simulator, in Fig. \ref{fig:output} we also show four hand-drawn images of arbitrary curves, which are given to the model to reconstruct based on the Fourier coefficients generated by the encoder. Though there are well-known methods for calculating the Fourier decomposition of a curve, this method can generalize to calculate the most likely result even given input that is outside the domain of the original function (a hand-drawn image vs. a vector of data or a plot in the original format generated by the simulator). 

\begin{figure}
  \centering
  \includegraphics[width=.9\textwidth]{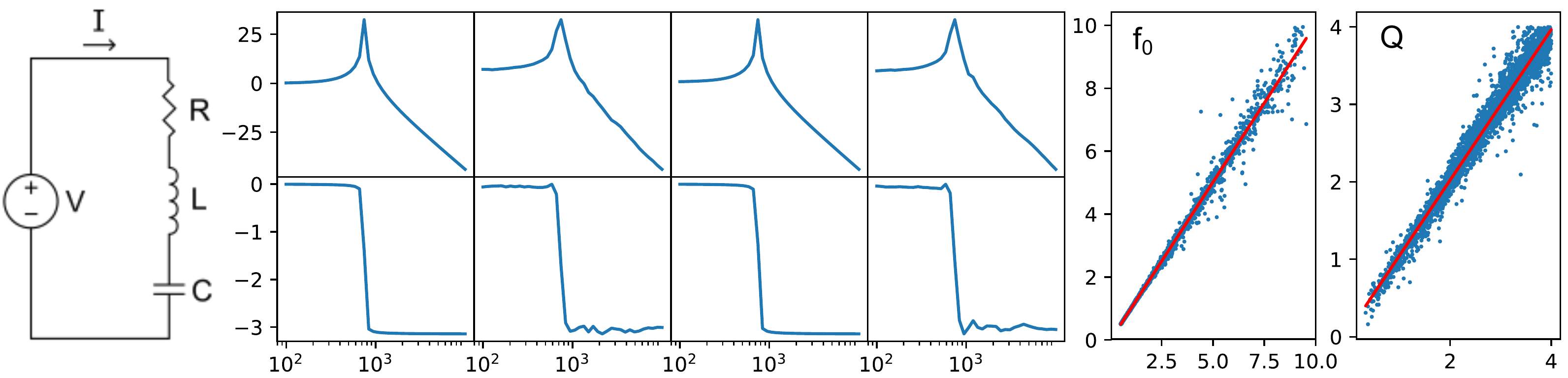}
  \caption{\textbf{Output of RLC model.} Left: schematic of the series RLC circuit. Center: comparison of decoder-generated and simulator-generated output gain in dB (top) and phase in radians (bottom). From left to right, the plots show simulator output $X$ on input $Z$, generator output $\bar{X}$ on the same input $Z$, the simulator output $X$ on encoder output $\bar{Z}$, and the generator output $\bar{X}$ on encoder output $\bar{Z}$. Right: correlation between $f_0$ (kHz) and $Q$ calculated from $Z$ vs from $V$, with calculated slopes of 1.00 and 0.97, and $R^2$ of 0.99 and 0.97 respectively. Data was limited to the range of validity of the model, $f_0$<10kHz and $Q$<4.}
  \label{fig:RLC}
\end{figure}

\paragraph{RLC circuit}
Finally, we present an example that is outside the domain of images altogether. Using the pySpice circuit simulator package (\cite{pyspice}), we simulate the series RLC circuit shown in Fig. \ref{fig:RLC}, with the simulator input being the resistance $R$, inductance $L$, and capacitance $C$, and the output being the gain and phase vs. frequency.
The inverse RLC simulator is also a one-to-many map, as the gain and phase are invariant with respect to a transformation of the form $R \mapsto R/x$, $L \mapsto L/x$, and $C \mapsto Cx$, where $x$ is any scaling factor. The relevant physical parameters are the resonant frequency $f_0 = 1/(2\pi \sqrt{LC})$ and quality factor $Q = (1/R)\sqrt{L/C}$.

\subsection{Disentanglement metrics}
\label{sec:metrics}





Given their joint and marginal distributions, mutual information (MI) between two discrete random variables $Z$ and $V$ is defined as 
\begin{align}
    \label{eq:mi}
    \text{MI}(Z,V) = \frac{1}{H(v)} \sum_z \sum_v p(z,v) \log \frac{p(z,v)}{p(z)p(v)}.
\end{align}

We can then define the mutual information matrix (MIM) as the element-wise MI between each element of $Z$ and $V$, and the mutual information gap (MIG) as defined in \citep{MIG} as the difference in MI between the two latent variables with highest MI for a given factor. Then one measure of disentanglement is the comparison between the mutual information between $Z$, the latent variable of the model, and $V$, the input of the simulator, with the ground truth being the mutual information between $V$ and itself.

For the ground truth MIM and MIG calculations, we take into account the one-to-many mappings of the inverse simulator. For example, if the polygon plotter sees a shape, any ordered permutation of the four points in the shape will generate the same shape, and will be a valid $Z$ for the encoder to learn. Therefore for simulators with one-to-many mappings, we calculate the ground truth mutual information as MIM($V$,$\bar{V}$) and MIG($V$,$\bar{V}$), where $\bar{V}$ represents $V$ under an invariance transformation: scaling for the RLC circuit and permutation for the polygon plotter.



\subsection{Results}

\begin{table}
  \caption{Mutual information gap score of \xxxVAE models.}
  \label{table:disentanglement}
  \centering
  \begin{tabular}{llll}
    \toprule
     Simulator     & MIG score & Ground truth MIG score & Ground truth correlations \\
    \midrule
      Fourier transform  &     0.76       & 1     & Independent \\
      dSprites  &           0.32   & 1 & Independent      \\
      Box plotter        &    0.56    & 0.82  & Correlated in sampling space         \\
             RLC circuit    &     0.13     & 0.07    & Invariant under scaling      \\
             Polygon plotter         &     0   & 0 & Invariant under permutation            \\
    \bottomrule
  \end{tabular}
\end{table}

\begin{figure}
  \centering
  \includegraphics[width=\textwidth]{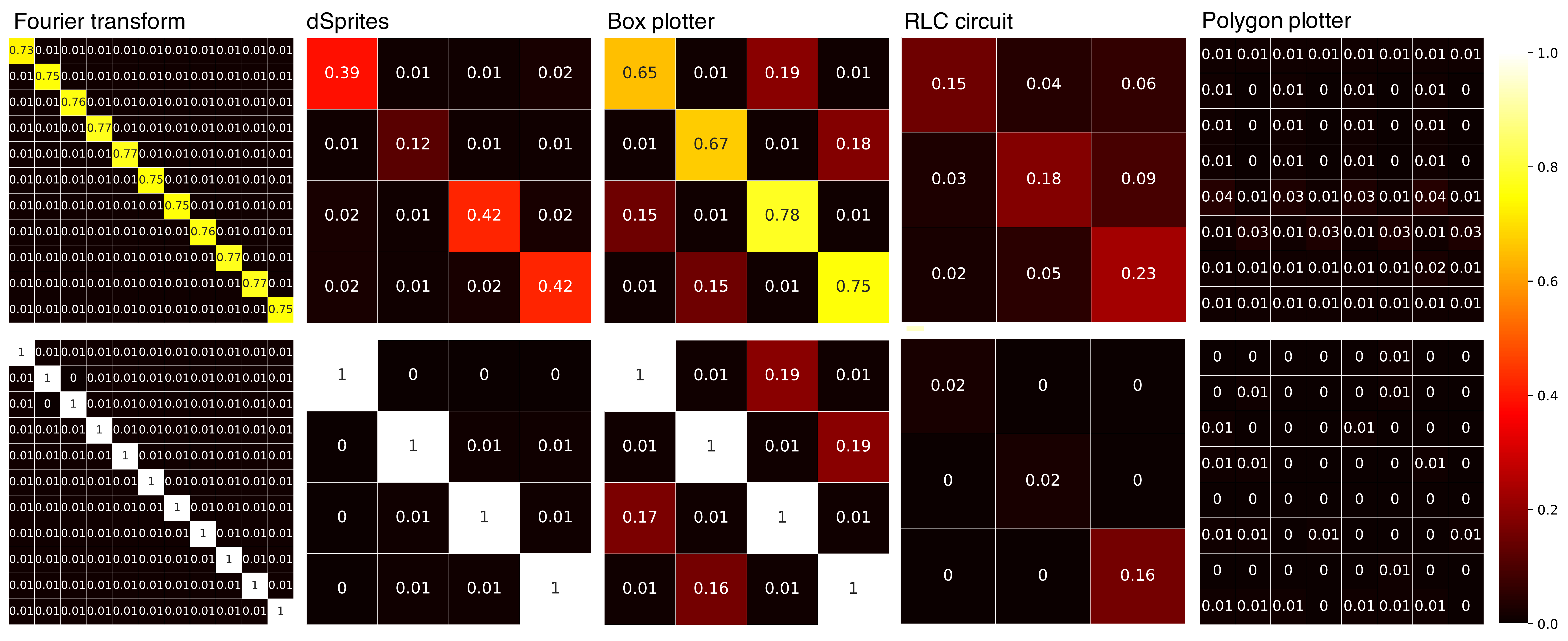}
  \caption{\textbf{Visualizations of mutual information matrices.} The top row is MIM($Z$,$V$), the mutual information of the model with the simulator, and the bottom row is the ground truth mutual information, MIM($V$,$V$) or MIM($V$,$\bar{V}$) as described in Sec. \ref{sec:metrics}.
}
  \label{fig:MI}
\end{figure}

For all models, MIG scores are given in Table \ref{table:disentanglement} and MIM visualizations are plotted in Fig. \ref{fig:MI}. In the \xxxVAE where each element of $Z$ should be identically matched with the corresponding element of $V$, the MIM matrix is square and carries an intuitive meaning: the diagonal elements specify how accurately the model has learned the simulator (how close $Z$ comes to exactly matching the simulated $V$), and the off-diagonal elements represent the correlations between the variables in the latent space, and therefore the degree of orthogonality. In contrast, the commonly used disentanglement metric MIG has the drawback of penalizing even a model which perfectly matches the intrinsic correlations between the ground truth factors. Additionally it combines the effects of representation accuracy and disentanglement (reductions in either will lower the MIG).

Even if the model does not perfectly learn the action of the simulator, due to insufficient depth, training time, suboptimal choice of architecture, etc, it is still possible to demonstrate a high degree of disentanglement using the MIM metric. For example, for the Fourier transform model, the MIG score is 0.76. However, we can see from the MIM that all of the weight is in the diagonal elements, and therefore in practice this is a fully disentangled representation. The missing mutual information is from insufficient model strength to match the simulator, and not in any correlations between the latent variables.

This similarly holds for the dSprites dataset - the latent variables are fully orthogonal, even though the representation accuracy is not complete. As calculated in \citep{DisentanglementLib}, MIG scores for a selection of models
on the dSprites dataset fall between a wide range of 0-0.4 depending on the choice of hyperparameters. Our model gives a MIG score of 0.32 on the (scale, rotation, $x$-position, $y$-position) factors of dSprites (treating the shape factor as noise, as in \citep{MIG}), which falls within this range.
But as can be seen from the full MIM in Fig. \ref{fig:MI}, the latent variables are fully disentangled, and the limit of mutual information is due to the quality of fit to the dataset rather than information mixing between latent variables. A different neural network architecture could be used to improve fit, or if a more accurate fit were prioritized over the matching of $Z$ with $V$, the latent space could be expanded with additional non-interpretable latent variables.

Note, however, that a lack of orthogonality also does not necessarily imply that the latent space is not interpretable. In the example of the 2D box plotter, the simulator was formulated with an inherent correlation between the variables of $V$: that the box must always remain fully within the specified plotting range, and therefore the width and height are correlated with the $x$ and $y$ coordinates respectively. Given this simple correlation built into the system, we can see that even in correlating $V$ with itself - as would be the case when $Z$ perfectly learns $V$ - there remains mutual information between the latent variables. By penalizing this correlation, as would be imposed for \betaVAE for example, the model would be required to deviate from the natural variables specified in the problem and would likely have a reduced representation accuracy. 

In cases such as the RLC circuit and polygon plotter, where the inverse simulator has a one-to-many mapping, both the MIM and MIG are uninformative. Though the latent variables $Z$ themselves are meaningful, as demonstrated by the quality of representation and transversals, they don't have a fixed relationship with the $V$ values from the simulator. Hence, the mutual information drives towards zero as the space for scaling and permutation expands. We raise a task for future work to identify a metric to quantify disentanglement and interpretability that can distinguish between representation accuracy and disentanglement, and which is effective in the case of many-to-one and one-to-many mappings.


\section{Discussion}
\label{sec:discussion}

In this paper we present \xxxVAE, a new approach for simulator-based training of variational autoencoders. Our approach is general and can hypothetically work with any simulator (assuming that the backbone network is expressive enough). We demonstrated its breadth by applying it to several domains, some of which are very different from visual scene understanding which was the focus of previous work on learning disentangled representations. A key aspect of our approach is to separate the training to two stages: a supervised stage where a generative model approximating the simulator is learned, and an unsupervised stage where an inference model approximating the inverse of the simulator is learned. It is interesting to note that typically generative models are associated with unsupervised learning while inference (discriminative) models are associated with supervision. Here we show that the opposite can yield powerful models.   

When considering the world as a complex simulator, our decoupled semi-supervised approach suggests interesting directions to explore in the context of machines vs. humans learning. A \xxxVAE model can be initialized in an unsupervised manner using a VAE or GAN approach (in our experiments we did not need this stage), followed by a supervised stage of training the generator and then an unsupervised one of training the encoder. Moving forward, both models can be continuously trained depending on the data that the system receives: when the observed data is coupled with labels (namely, values of latent variables) the supervised generator training kicks in, and when it is not, the unsupervised encoder training kicks in. When babies sense the world, they similarly receive both labeled and unlabaled data: most of the time they observe the world without supervision, but every now and then someone points to an object saying 'this is a red ball'. It is therefore intriguing to explore whether alternating between supervised and unsupervised learning of approximate simulation and its inverse respectively, are at work in human learning.    

An interesting direction for future research is to explore compositionality. As opposed to unsupervised approaches such as InfoGAN or \betaVAE, our approach can learn exactly the latent variables of interest for downstream applications. For example consider the problem of learning to multiply hand written numbers. One can first learn models that generate and recognize hand written digits by approximating a digits graphics engine. Then given a multiplication simulator, one can learn models that multiply and factor hand written numbers up to 81 (9X9) and use the digits generator to generate the 'hand written' result. In order to learn in general how to multiply multi-digit numbers, causality is key \citep{Lake:2016}. Namely, fine grained simulators of the components of the multiplication algorithm will have to be learned (note that this is how children learn to do it at school). This idea, with a different learning technique, was explored in the context of hand written characters \citep{Lake:2015}. In future work we plan to explore compositionality and causality as well as applications of our approach to more complex simulators such as photo-realistic graphics engines.

\bibliography{neurips} 
\bibliographystyle{icml2017}

\end{document}